\def\year{2018}\relax
\documentclass[letterpaper]{article} 
\usepackage{aaai18}  
\usepackage{times}  
\usepackage{helvet}  
\usepackage{courier}  
\usepackage{url}  
\usepackage{graphicx}  
\usepackage{algorithm}
\usepackage{algorithmicx}
\usepackage{algpseudocode}
\usepackage{subfigure}

\def\ie{\emph{i.e.}}

\def\etc{\emph{etc}}
\def\etal{{\em et al.~}}
\graphicspath{{figures/}}

\nocopyright

\usepackage{fancyhdr}

\begin{document}
%
\title{FLIC: Fast Linear Iterative Clustering with Active Search}
\author{Jiaxing Zhao$^1$ \quad Bo Ren$^1$\thanks{Bo Ren is the corresponding author: rb@nankai.edu.cn.} \quad Qibin Hou$^1$ 
  \quad Ming-Ming Cheng$^1$ \quad Paul L. Rosin$^2$\\
$^1$Nankai University \quad $^2$Cardiff University}
\maketitle

\thispagestyle{fancy}
\fancyhead{}
\fancyhead[L]{AAAI 2018}
\renewcommand{\headrulewidth}{0pt}

\begin{abstract}
In this paper, we reconsider the clustering problem for image over-segmentation from a new perspective.
We propose a novel search algorithm named ``active search''  which 
explicitly considers neighboring continuity. Based on this search 
method, we design a back-and-forth traversal strategy and a 
``joint'' assignment and update step to speed up the algorithm. 
Compared to earlier works, such as Simple Linear Iterative Clustering 
(SLIC) and its follow-ups, who use fixed search regions and perform the assignment
and the update step separately, our novel scheme reduces the 
number of iterations required for convergence,
and also improves the boundary sensitivity of the over-segmentation results.
Extensive evaluations on the Berkeley segmentation benchmark 
verify that our method outperforms competing methods under various
evaluation metrics.
In particular, lowest time cost is reported among existing 
methods (approximately $30\mbox{ fps}$ for a $481 \times 321$ image
on a single CPU core).
To facilitate the development of over-segmentation, the code will be
\emph{publicly available}.
\end{abstract}

\section{Introduction}
Superpixels, generated by image over-segmentation, take the place
of pixels to become the fundamental units in various computer vision 
tasks, including image segmentation \cite{cheng2016hfs}, image classification
\cite{wang2013image}, 3D reconstruction \cite{hoiem2005automatic},
object tracking \cite{wang2011superpixel}, \etc.
Such a technique can greatly reduce computational complexity,
avoid under-segmentation, and reduce the influence caused by noise.
Therefore, how to generate superpixels with high efficiency plays an important
role in many vision and image processing applications.

Generating superpixels has been an important research issue,
and a group of classical methods have been developed,
including FH \cite{felzenszwalb2004efficient},
Mean Shift \cite{comaniciu2002mean},
Watershed \cite{vincent1991watersheds}, \etc.
The lack of compactness and the irregularity of superpixels restrict their
applications, especially when contrast is
poor or shadows are present.
To solve the above-mentioned problems,
Shi and Malik proposed Normalized Cuts (NC)
\cite{shi2000normalized} that generated compact superpixels.
However, this method does not adhere to image boundaries very well,
and the complexity is high.
GraphCut \cite{boykov2001fast,veksler2010superpixels} 
regarded the segmentation problem as an energy optimization
process.
It solved the compactness problem by using min-cut/max-flow 
algorithms \cite{boykov2004experimental,kolmogorov2004energy}, but
their parameters are hard to control.
Turbopixel \cite{levinshtein2009turbopixels} is another method
that is proposed to solve the compactness problem.
However, the inefficiency of the underlying level-set method
\cite{osher1988fronts} restricts its applications.
Bergh \etal~\shortcite{van2012seeds} proposed an energy-driven algorithm
SEEDS whose results adhered to the boundaries
well, but unfortunately it suffers from irregularity and the 
number of superpixels is uncertain.
ERS \cite{liu2011entropy}, although it
performs well on the Berkeley segmentation benchmark, has a
high computational cost that limits its practical use.

Achanta \etal \shortcite{achanta2012slic} 
proposed a linear clustering based algorithm SLIC.
It generates superpixels based on Lloyd's algorithm
\cite{lloyd1982least} (also known as Voronoi iteration or $k$-means).
In the assignment step of SLIC, as a key point to speed up the algorithm, each pixel
$p$ is associated with those cluster seeds whose search regions overlap its location.
Such a strategy is also adopted by most subsequent works based on SLIC.
SLIC is widely used in various applications \cite{wang2011superpixel}
because of its high efficiency and good performance.
Inspired by SLIC, Wang \etal \shortcite{wang2013structure} implemented an algorithm SSS that considered the structural information within images.
It uses the geodesic distance \cite{peyre2010geodesic} computed
by the geometric flows instead of the simple Euclidean distance.
However, efficiency is poor because of the bottleneck
caused by the high computational cost of measuring geodesic distances.
Very recently, Liu \etal proposed Manifold SLIC~\shortcite{liu2016manifold} that
generated content-sensitive superpixels by computing Centroidal Voronoi
Tessellation (CVT) \cite{du1999centroidal} in a special feature space.
Such an advanced technique makes it much faster than SSS but still slower
than SLIC owing to the cost of its mapping, splitting and merging processes.
From the aforementioned descriptions, we see that the above-mentioned 
methods improve the results by either using more complicated
distance measurements or providing more suitable transformations 
of the feature space.
However, the assignment and update steps within these methods are 
performed separately, leading to low convergence rate.

In this paper, we consider the over-segmentation problem from a new perspective.
Each pixel in our algorithm is allowed to actively search which superpixel
it should belong to, according to its neighboring pixels as shown in Figure \ref{fig:assignment}.
In the meantime, the seeds of the superpixels can be 
adaptively changed during this process, 
which allows our assignment and update steps to be performed jointly.
This property enables our approach to converge rapidly.
To sum up, our main advantages are:
\begin{itemize}
\item Our algorithm features good awareness of neighboring-pixel 
continuity, and produces results with good boundary sensitivity 
regardless of image complexity and contrast.
\item Our algorithm performs the assignment step and the update step
in a joint manner, and has a high convergence rate as well as the lowest time cost
among all superpixel segmentation approaches.
Experiments show that our approach is able to converge in two scan loops,
with better performance measured under a variety of evaluation metrics
on the Berkeley segmentation benchmark.
\end{itemize}

\section{Preliminaries}\label{sec:Pre}

Before introducing our approach that allows adaptive search regions and joint assignment and update steps, we first briefly recap the standard previous scheme with fixed search regions and separate steps. A typical one is the SLIC algorithm who improves Lloyd's algorithm, reducing the time
complexity from $O(KN)$ to $O(N)$, where $K$ is the number of the superpixels and $N$ is the number of pixels.

\begin{figure}[t]
	\centering
	\includegraphics[width = \linewidth]{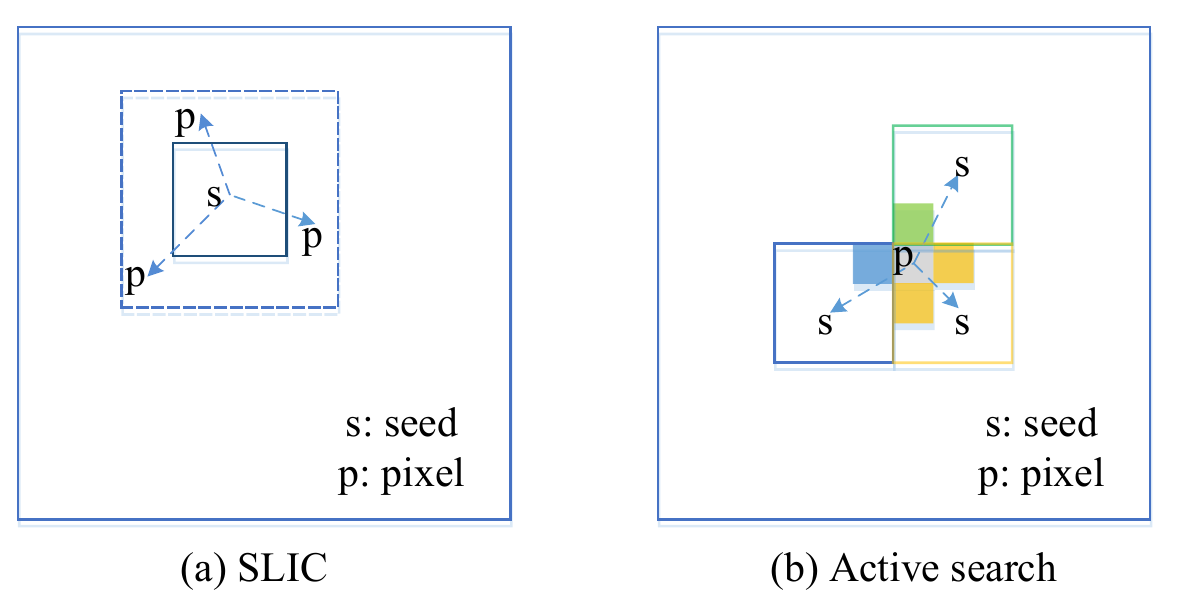}
	\caption{(a) ~The search method used in SLIC. Each seed
         only searches a limited region to reduce computation complexity.
         (b)~Our proposed active search.
		Each pixel is able to decide its own label by searching its surroundings.
    }
    \label{fig:assignment}
\end{figure}

Let $\{I_i\}_{i=1}^{N}$ be a color image, where
$I_i$ represents the corresponding variable of each pixel.
Given a set of evenly distributed seeds $\{S_k\}_{k=1}^{K}$,
SLIC simplifies the Lloyd's algorithm to get the
Centroidal Voronoi Tessellation (CVT)~\cite{du1999centroidal}
that will be introduced in Section \ref{3.3}.
In the assignment step, each pixel $I_i$ is associated with those
cluster seeds whose search regions overlap its location as shown in
Figure \ref{fig:assignment}(a).
The area of a search region can be denoted by $2T \times 2T$,
where $T=\sqrt{N/K}$.
Specifically, SLIC considers $I_i$ to lie in a five dimensional space that
contains a three dimensional CIELAB color space $(l_i,a_i,b_i)$ and a
two dimensional spatial space $(x_i,y_i)$.
SLIC measures the distance between two points using a weighted Euclidean distance,
which can be computed by
\begin{equation}\label{1}
D(I_i,I_j)= \sqrt{d_c^2 + \bigg(\frac{d_s*m}{N_s}\bigg)^2},
\end{equation}
where $m$ is a variable that controls the weight of
the spatial term, and $N_s=T$.
Variables $d_s$ and $d_c$ are respectively the spatial and color
distances, which can be expressed as
\begin{equation}
d_s=\sqrt{(x_i-x_j)^2+(y_i-y_j)^2},
\end{equation}
and
\begin{equation}
d_c = \sqrt{(l_i-l_j)^2+(a_i-a_j)^2+(b_i-b_j)^2}.
\end{equation}
In the update step, SLIC recomputes the center of each superpixel and moves
the seeds to these new centers.
Then it obtains the over-segmentation results by iteratively performing the
assignment and update steps.

The follow-up works of SLIC also use a similar procedure as SLIC.
They improve the performance of SLIC using better distance measures or
more suitable transformation function
between color space and spatial space.
However, in these algorithms, each search region is fixed in the assignment
step of a single loop, and the relationship among
neighboring pixels is largely ignored when allocating pixels to superpixels.
Separately performing the assignment step and the update step also leads to
a delayed feedback of pixel label change.

\section{The Proposed Approach}
\label{sec:method}

Since superpixels normally serve as the first step of other
vision related applications, how to generate superpixels with good 
boundaries and very fast speed is a crucial problem.
Here, unlike previous algorithms~\cite{achanta2012slic,liu2016manifold}, we consider
this problem from a new aspect, in which only the surrounding pixels
are considered for determining the label of the current pixel. Each pixel actively selects which superpixel it 
should belong to in a back-and-forth order to provide better estimation of over-segmentation regions.
Moreover, the assignment step and the update step are performed jointly. Very few iterations are required for our approach to reach convergence.
An overview of our algorithm can be found in Alg. \ref{alg1}.

\begin{algorithm}[t]
	\caption{FLIC}
	\begin{algorithmic}
	  \Require Image $I$ with $N$ pixels, the desired number of superpixels $K$,
        the maximal iteration numbers $itr_{max}$ 
        and the spatial distance weight $m$.
	  \Ensure $K$ superpixels
		\State Divide the original image into regular grids $\{G_k\}_{k=1}^{K}$
		\State \hspace{.1in}	  with step length $\upsilon = \sqrt{N/K}$.
		\State Initialize labels $\{L_k\}_{k=1}^{K}$ for pixels according to their
		\State \hspace{.1in}locations.
		\State Move each seed to the lowest gradient position in 3$\times$3
		\State \hspace{.1in}neighborhoods.
		\State Initialize seeds $\{S_k\}_{k=1}^{K}$.
		\State Regard pixels sharing the same label as a superpixel $\zeta$.
		\State Initialize distance $d_{(i)} = \infty $ for each pixel and $itr = 0$.
		\While {itr $<$ $itr_{max}$}
		  \For {each superpixel $\zeta_k$}
			\State {Use back-and-forth scan to traverse superpixel $\zeta_k$
			\State \hspace{.05in} to get the pixels  processing sequence 
			  (\S \ref{order}).}
			\For {each pixel $I_i$ in the sequence}
			\State Set $d_{(i)} = D(I_i,S_{L_i}) $  by  Eqn. \ref{1}
			\For {$I_j$ in the four-neighborhood of $I_i$}
			\If {$L_j$ $ \neq $ $L_{i}$}
			\State Compute
			    $D=D(I_i,S_{L_j})$. (Eqn. \ref{1})
			\If {$ D < d_{(i)}$}
			\State $d_{(i)}= D$; ~$L_i=L_{j}$.
			\EndIf
			\EndIf
		  \EndFor			
           \If {$L_i$ is changed to $L_j$}
             \State Use Eqn. \ref{10} to update $\zeta_{L_i}$;
             \State Use Eqn. \ref{20} to update $\zeta_{L_j}$;
             \State Update the bounding box of $\zeta_{L_j}$ (\S \ref{3.3}).
           \EndIf
          \EndFor
	    \EndFor
		\State itr++;
	  \EndWhile
	\end{algorithmic}
	\label{alg1}
\end{algorithm}

\subsection{Problem Setup}

Given the desired number of superpixels $K$
and an input image $I = \{I_i\}_{i=1}^{N}$,
where $N$ is the number of pixels,
our goal is to produce a series of disjoint
small regions (or superpixels).
Following most previous works \cite{achanta2012slic}, 
the original RGB color space is transformed to the CIELAB color space  
(which has been proven useful).
Thus, each pixel $I_i$ in an image $I$ can be represented in a five
dimensional space,
\begin{equation}\label{2}
	I_i = (l_i,a_i,b_i,x_i,y_i).
\end{equation}
We first divide the original image into a regular grid containing $K$ elements $\{G_k\}_{k=1}^{K}$
with step length $\upsilon = \sqrt{N/K}$ as in \cite{achanta2012slic}, and the
initial label for each pixel $I_i$ is assigned as:
\begin{equation}\label{3}
	L_i = k,~\textit{if}  ~I_i \in G_k.
\end{equation}
We initialize the seed $S_k$ in $G_k$ as the centroid.
Therefore, $S_k$ can also be defined in the same five dimensional space
\begin{equation}\label{denote_seed}
S_k = \{l_k,a_k,b_k,x_k,y_k\}.
\end{equation}
	
\subsection{Label Decision} \label{sec:label_decision}

\begin{figure*}[t]
    \centering
 	\includegraphics[width = 0.95\linewidth]{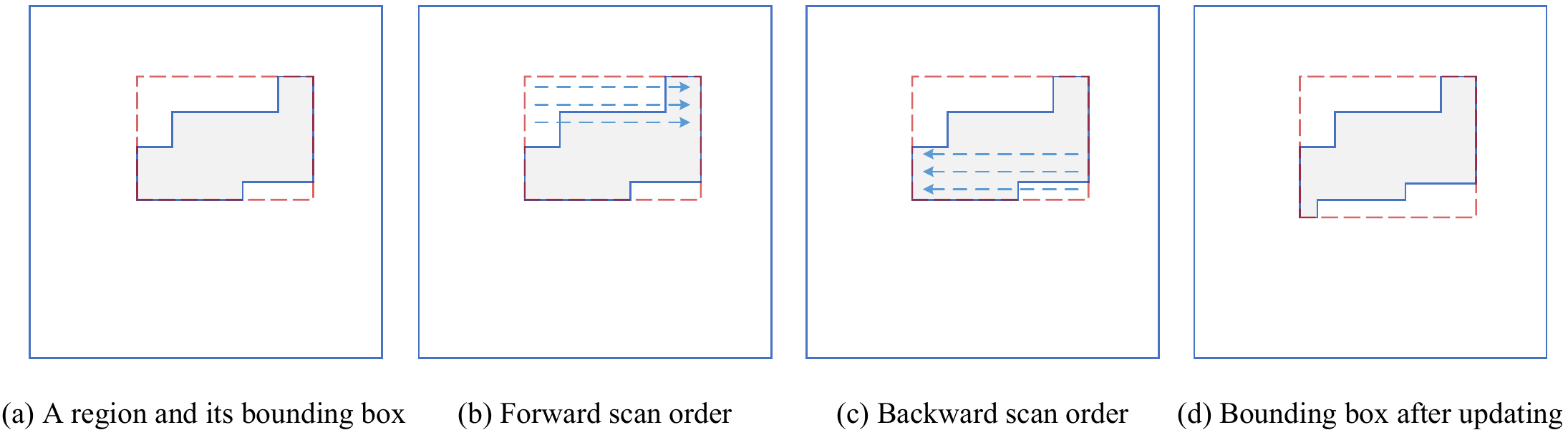}
 	\caption{Illustration of our scanning order for each superpixel.
        We use gray regions enclosed by blue lines to represent superpixels
        and use red dashed rectangles to denote their corresponding bounding
        boxes. As shown in (b) and (c), we first scan the bounding box
        from left to right and top to bottom and then in the opposite direction.
        The shape of each superpixel might change, and so we
        update the bounding box if this occurs, as in (d), and leave it
        unchanged if there are no changes to the superpixel shape.
 	}\label{scan_normal}
\end{figure*}

In most natural images adjacent pixels tend to share the same labels, 
\ie~neighboring pixels have natural continuity.
Thus, we propose an active search method that is
able to leverage as much of this a priori information as possible.
In our method, unlike most previous works \cite{achanta2012slic,liu2016manifold},
the label of the current pixel is only determined by its neighbors.
We will compute the distances between the current pixel and the seeds of its
four or eight adjacent pixels.
Figure \ref{fig:assignment} provides a more intuitive illustration.
Specifically, for a pixel $I_i$, our assignment principle is
\begin{equation} \label{assign_eqn}
L_i =  \mathop {argmin}_{L_j}D(I_i, S_{L_j}), I_j \in A_i,
\end{equation}
where $A_i$ consists of $I_i$ and its four neighboring pixels, 
$S_{L_j}$ is $I_j$'s corresponding superpixel seed. 
We use Eqn. \ref{1} to measure the distance $D(I_i,S_{L_j})$.

Since each pixel can only be assigned to a superpixel containing at least one of 
its neighbors, the local pixel continuity has a stronger effect in the proposed 
strategy, allowing each pixel to actively assign itself to one of its surrounding 
closely connected superpixel regions. 
The advantages of such a strategy are obvious.
First, the nearby assignment principle can avoid the occurrence of
too many isolated regions, indirectly preserving the desired number of superpixels.
Second, such an assignment operation is not limited by a fixed range in space,
resulting in better boundary adherence despite some irregular shapes of those
superpixels with very complicated content.
Furthermore, in the assignment process, the superpixel centers are also 
self-adaptively modified, leading to faster convergence.
Detailed demonstration and analysis can be found in Section \ref{sec:alg_analysis}.
It is worth mentioning that the neighbors of the internal pixels in a 
superpixels normally share the same labels, so it is unnecessary to process
them any more.
This fact allows us to process each superpixel extremely quickly.

\subsection{Traversal Order} \label{order}

The traversal order plays a very important role in our approach in that
an appropriate scanning order may lead to a visually better segmentation.
As demonstrated in Section \ref{sec:label_decision}, the label of each pixel only depends
on the seeds of its surrounding pixels.
This indicates that, in a superpixel, the label of the current pixel is directly or
indirectly related to those pixels that have already been dealt with.
To better take advantage of this avalanche effect, we adopt a back-and-forth
traversal order as in PatchMatch~\cite{barnes2009patchmatch},
in which the pixels that are processed later will benefit from the previously processed pixels.
Figure \ref{scan_normal} makes this process clear.
In the forward pass, the label decision of each pixel considers the information
from the top surrounding pixels of the superpixel, and similarly, the backward pass
will provide the information from the bottom surrounding pixels of the superpixel.
With such a scanning order, all the surrounding information can be taken into
consideration, yielding better segments.

Considering that an arbitrary superpixel might have an irregular shape instead
of a simple rectangle or square, we actually use a simplified strategy to traverse
the whole superpixel.
For any superpixel, we first find a minimum bounding box within which all its 
pixels are enclosed, as shown in Figure \ref{scan_normal}.
We then perform the scanning process for all the pixels in the corresponding
minimum bounding box and only deal with those pixels that are within the
superpixel.
	
\subsection{Joint assignment and update step} \label{3.3}
A common phenomenon in existing methods, such as SLIC \cite{achanta2012slic},
is that the assignment step and the update step are performed separately,
leading to delayed feedback from pixel label changes to superpixel seeds.
An obvious problem of such a strategy is that many (normally more than five) iterations 
are required which becomes the bottleneck of fast convergence.
In our approach, based on the assignment principle Eqn. \ref{assign_eqn}, we design a
``joint'' assignment and update strategy which 
operate these two steps at a finer granularity. The approximately 
joint step is able to adjust the
superpixel seed center position on the fly, drastically
reducing the number of iterations needed for convergence.
Since most clustering-based superpixel methods use the Centroidal Voronoi Tessellation
(CVT), we will briefly introduce the CVT first and then describe our method.

Let $S= \{S_k\}_{k=1}^{K}$ be the set of seeds in the image, where $K$ is the
expected number of superpixels. 
The Voronoi cell $\mathcal{V}_{(S_k)}$ of a seed $S_k$ is denoted by:
\begin{equation} \label{4}
\mathcal{V}_{S_k}=\{ I_i \in I~|~d(I_i,S_k) \leq d(I_i,S_j), \forall S_j\in S\},
\end{equation}
where $d(I_i,S_k) $ is an arbitrary distance measure from pixel $I_i$ to the 
seed $S_k$.
The Voronoi Diagram $\mathcal{V}_D (S)$ is defined by
\begin{equation}\label{5}
\mathcal{V}_D (S) = \{\mathcal{V}_{S_k} \ne \phi ~|~ \forall S_k \in S \}.
\end{equation}
A CVT is then defined as a Voronoi Diagram whose generator point of each Voronoi 
cell is also its center of mass. 
As mentioned above, the traditional CVT is usually obtained by heuristic algorithms, 
such as Lloyd's algorithm, iteratively performing updates after each assignment step until
convergence is reached.

In our approach, on account of our novel label decision strategy as shown in Eqn. \ref{assign_eqn}, we are able to jointly perform the update step and the assignment step instead of separately.
More specifically, after pixel $I_i$ is processed, if its label is changed to,
for instance, $L_j$, we immediately update the current seed $S_{L_i}$ using the following equation:
\begin{equation}\label{10}
S_{L_i} = \frac{S_{L_i} *  \left| \zeta_{L_i} \right| - I_i} {\left| \zeta_{L_i} \right| - 1},
\end{equation}
where $\left|\zeta_{L_i}\right|$ is the number of pixels in superpixel $\zeta_{L_i}$,
and update $S_{L_j}$ using the following equation
\begin{equation}\label{20}
S_{L_j} = \frac{S_{L_j} * \left| \zeta_{L_j} \right| + I_i} {\left| \zeta_{L_j} \right| + 1}.
\end{equation}
The bounding box of $\zeta_{L_j}$ is also updated thereafter.

It is noteworthy to mention that the above updates only contain very simple
arithmetic operations and hence can be performed very efficiently.
Such an immediate update will help later pixels make a better choice during assignment, leading to better convergence.
Figure \ref{fig:jointly} shows the convergence speed of our approach.

\section{Experiments} \label{sec:result}

\begin{figure*}
	\centering
    \subfigure[BR curves]{\label{Boundary recall}
    \includegraphics[width = 0.24\linewidth]{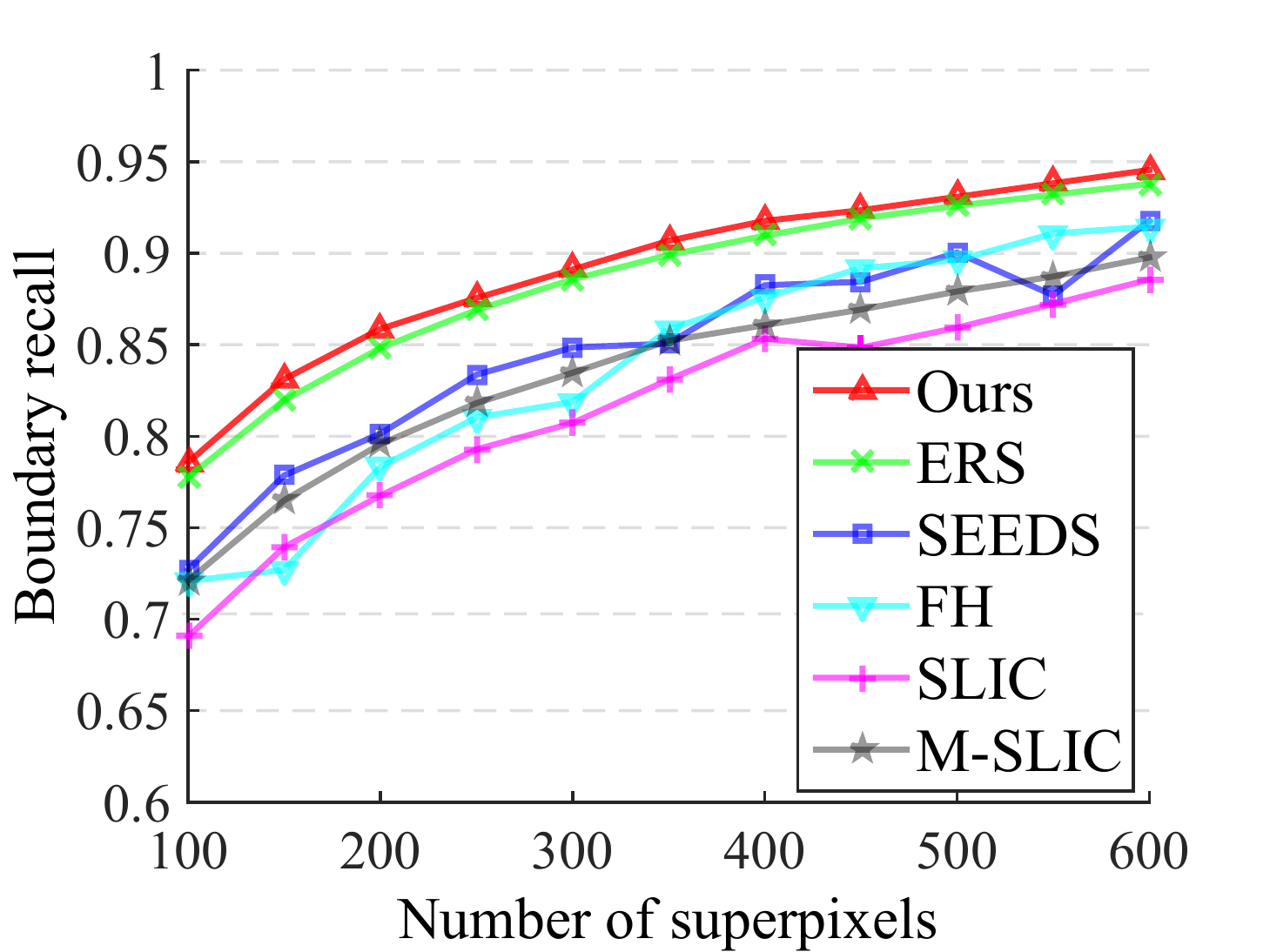}}
    \subfigure[UE trade-off]{\label{Undersegment error}
    \includegraphics[width = 0.24\linewidth]{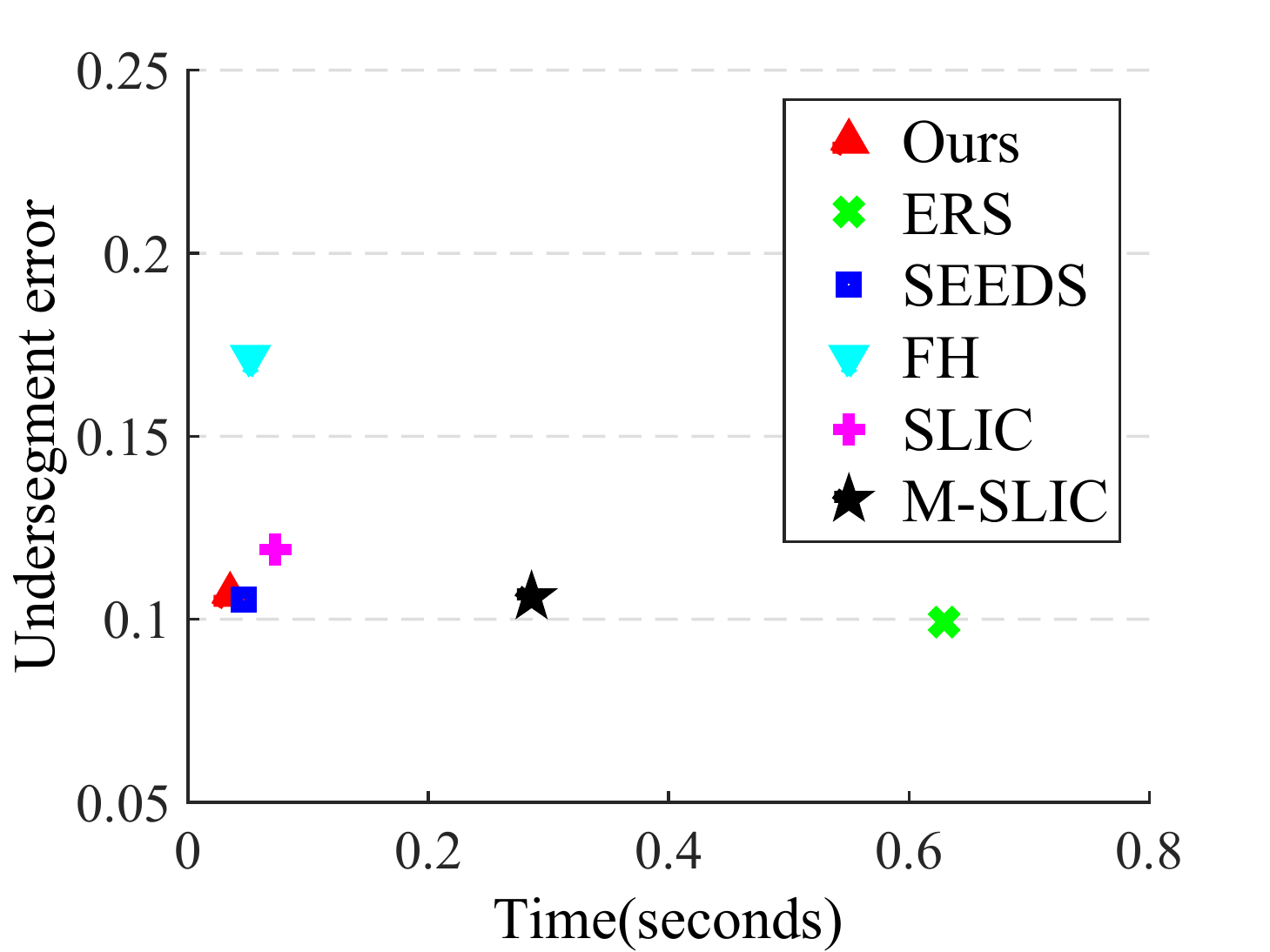}}
    \subfigure[ASA trade-off]{\label{fig:SegAcc}
    \includegraphics[width = 0.24\linewidth]{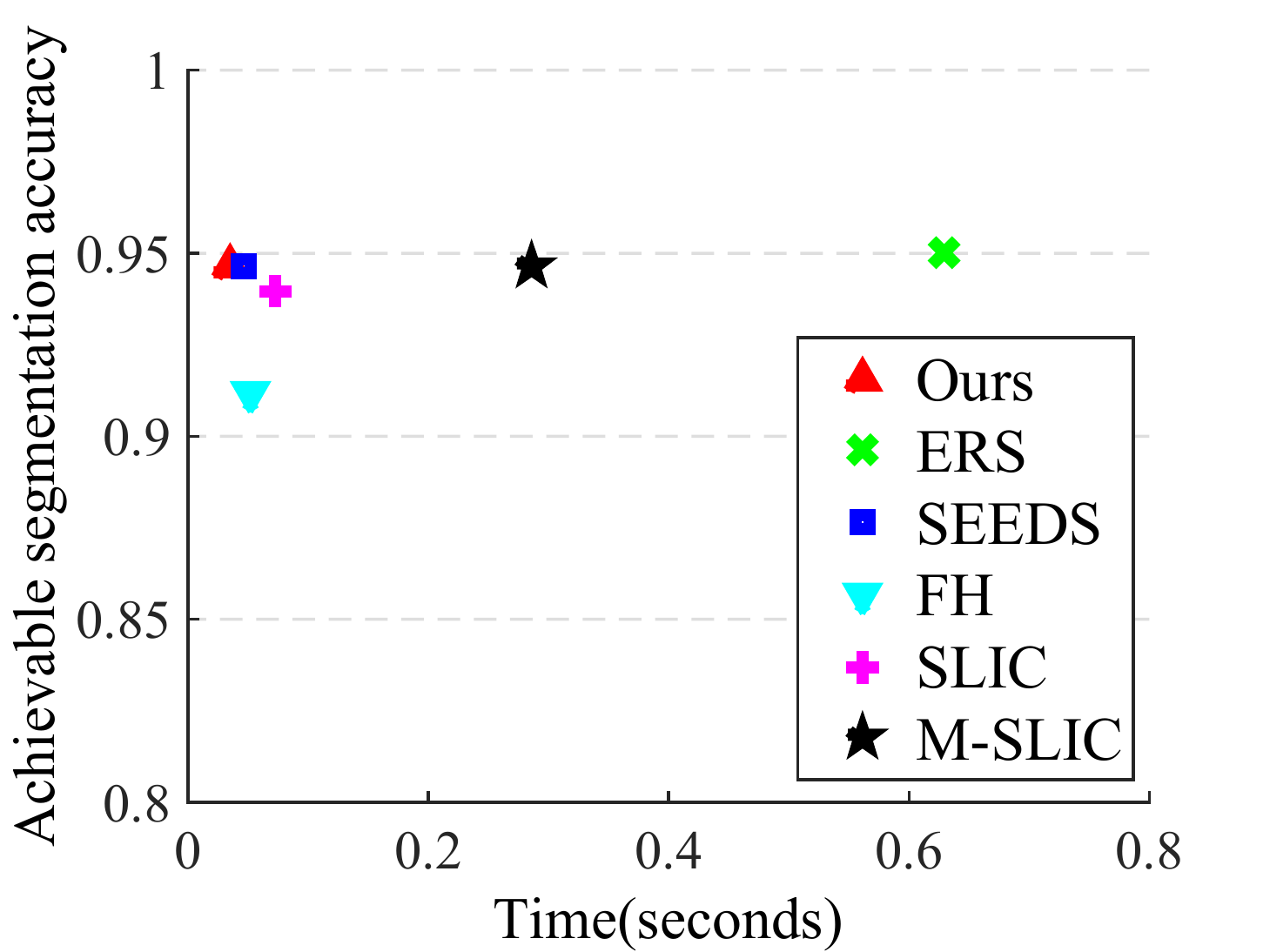}}
    \subfigure[BR trade-off]{\label{Time cost}
    \includegraphics[width = 0.24\linewidth]{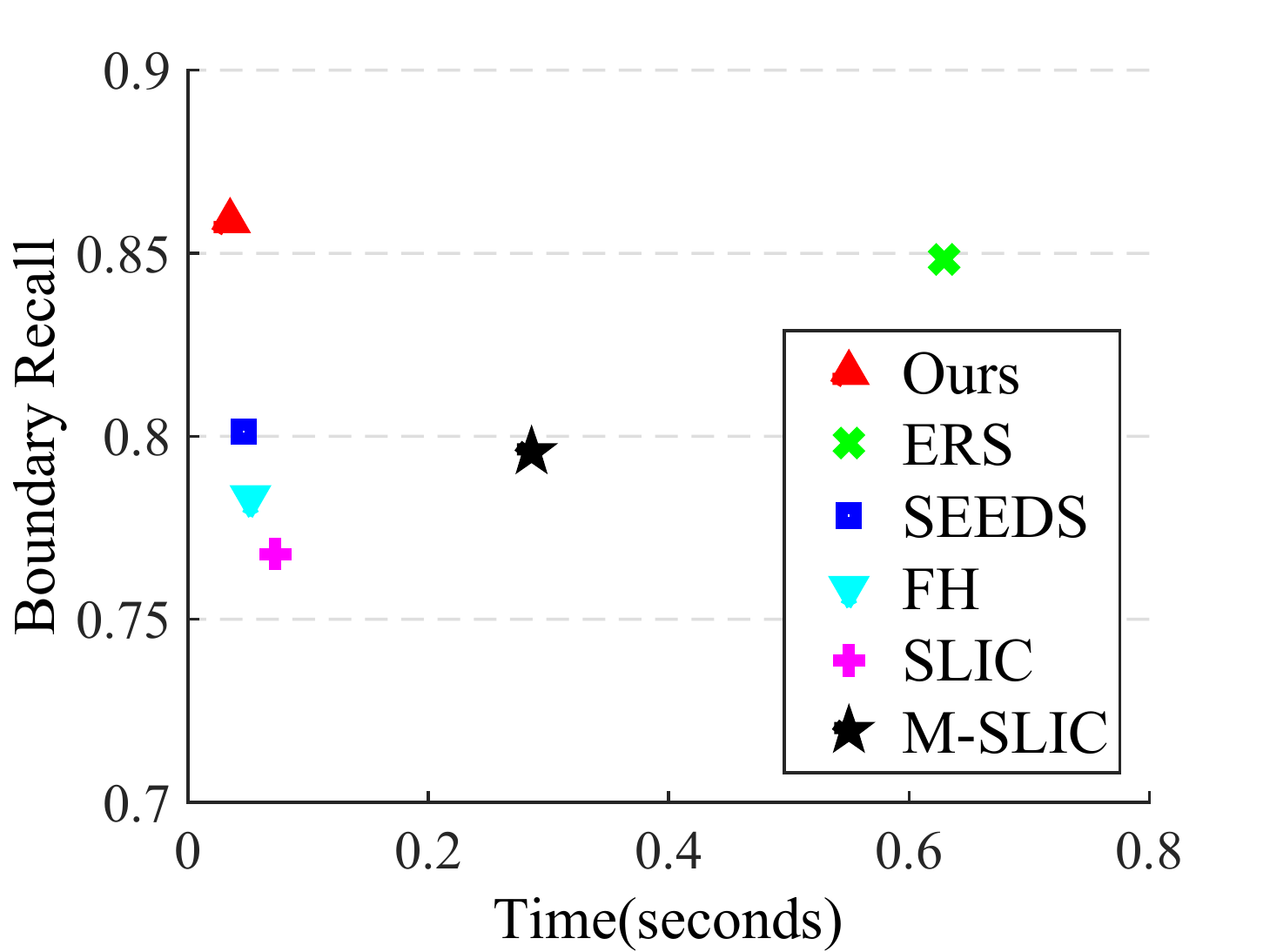}}
	\caption{Comparisons between existing state-of-the-art methods and 
    our approach (FLIC) on the BSDS500 benchmark.
    In (b)-(d), $K$ is fixed to $200$ to demonstrate our best trade-off between performance and efficiency between competing methods.
    As can be seen, our strategy significantly outperforms methods that have similar time cost in boundary recall. At least competitive results are also achieved
    compared to slower methods (e.g. the state-of-the-art method ERS \cite{liu2011entropy}) 
    according to all the evaluation metrics, but at an order of magnitude faster speed.}
    \label{evaluation}
\end{figure*}

\begin{figure}[t]
	\centering
    \subfigure[BR]{\label{Boundary recall_total}
    \includegraphics[width = 0.48\linewidth]{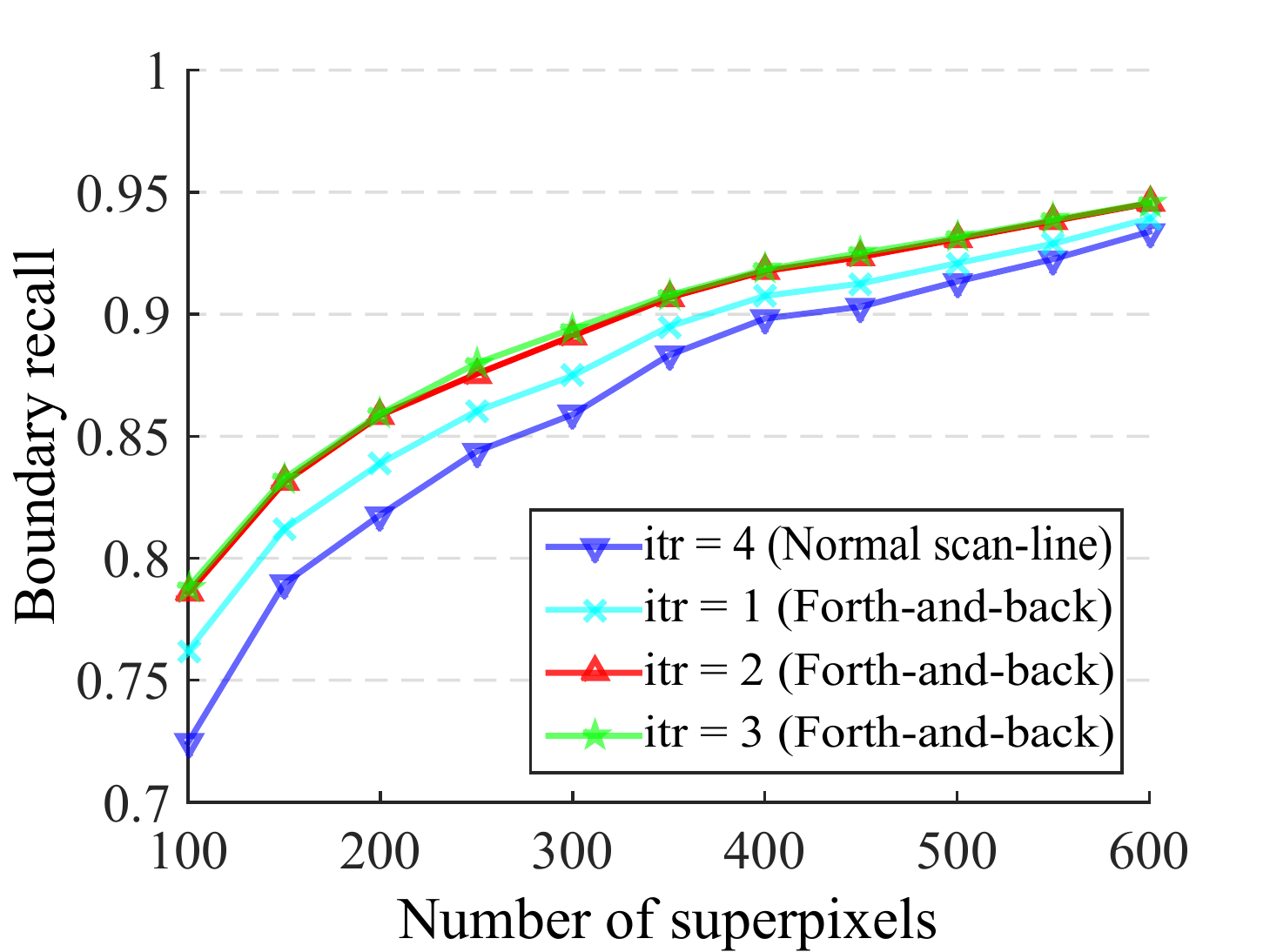}}
	\subfigure[Time Cost]{\label{Time_total}
    \includegraphics[width = 0.48\linewidth]{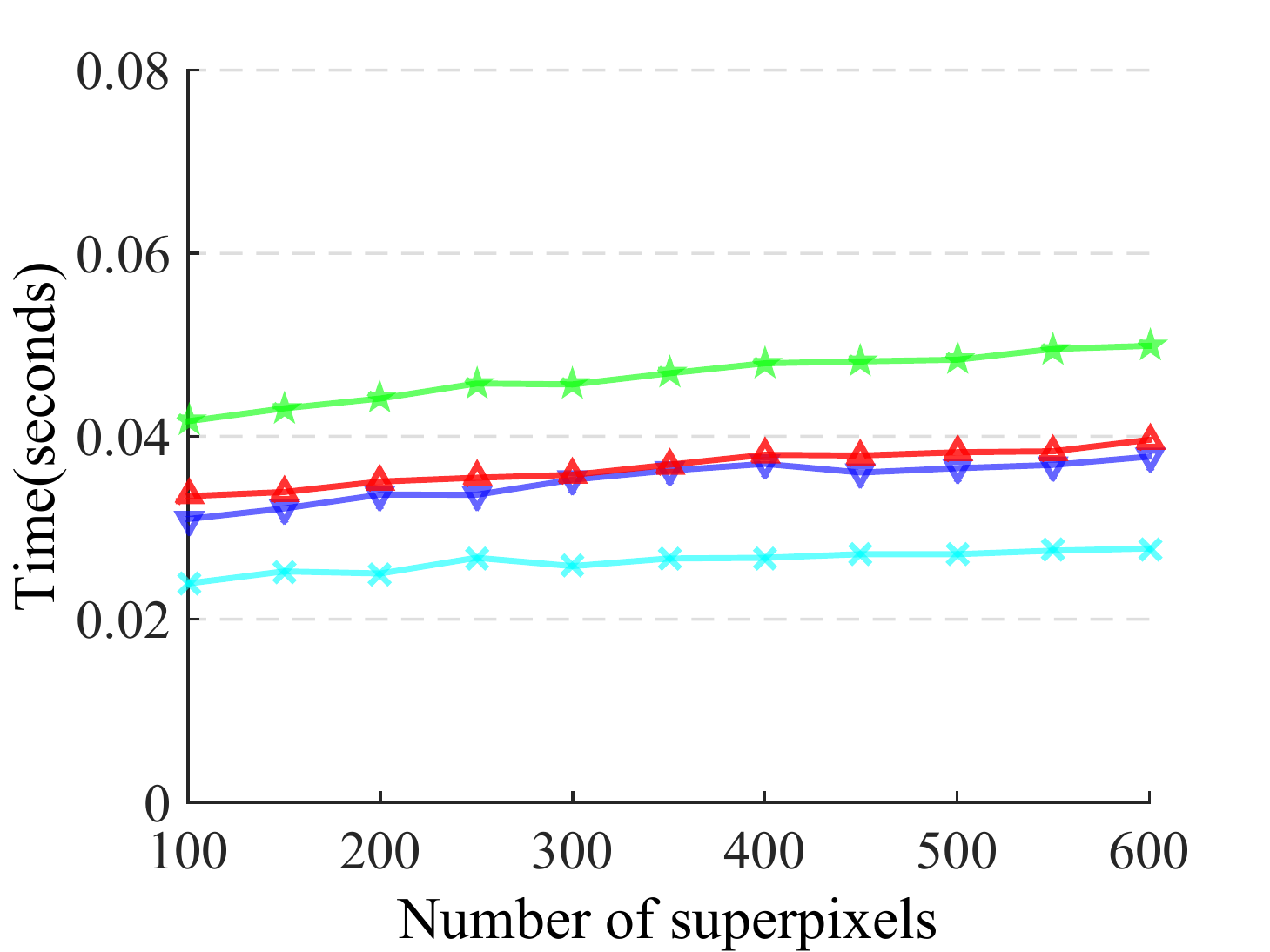}}
	\caption{Part of sensitivity analysis under the standard evaluation
    metrics and time cost.}\label{scan_order}
\end{figure}

Our method is implemented in C++ and runs on a PC with an Intel
Core i7-4790K CPU with 4.0GHz, 32GB RAM, and 64 bit operating system. 
We compare our method with many previous and current state-of-the-art works,
including FH~\cite{felzenszwalb2004efficient}, SLIC~\cite{achanta2012slic}, 
Manifold SLIC~\cite{liu2016manifold}, SEEDS~\cite{van2012seeds}, and
ERS~\cite{liu2011entropy} on the BSDS500 benchmark, using the evaluation
methods proposed in \cite{arbelaez2011contour,stutz2014superpixel}.
Note that the source codes used in evaluation of the above works may be of different versions, and we find this leads to performance difference from the original reports when a different implementation of evaluation code is applied.
To give a fair comparison, we uniformly use publicly available source code \cite{arbelaez2011contour,stutz2014superpixel} for all the methods.
As in previous research in the literature \cite{liu2016manifold,wang2013structure}, we evaluate all algorithms on 200 randomly selected images of resolution 
$481 \times 321$ from the Berkeley dataset.
	
\subsection{Parameters}
In our approach, three parameters need to be set. The first one is the number of superpixels $K$. 
One of the common advantages of clustering-based algorithms is that
the expected number of superpixels can be directly obtained by setting the 
clustering parameter $K$. 
The second one is the spatial distance weight $m$. 
Parameter $m$ has a large effect on the smoothness and compactness of superpixels. 
We shall show that our performance will increase as $m$ decreases.
However, a small $m$ can also lead to the irregularity of superpixels.
To achieve a good trade-off between compactness and performance, in the following
experiments, we set $m = 5$ as default.
The last parameter is the number of iterations $itr$. 
Here we set $itr = 2$ in default to get the balance between time cost and 
performance.
What should be stressed out here is that to compare with other methods \emph{in a fair way}, for each method we optimize its parameters to maximize the recall value computed on the BSDS500 benchmark.

\subsection{Comparison with Existing Methods} \label{4.2}
Our approach outperforms previous methods that have similar computational efficiency , and achieve at least comparable results compared to slower algorithms with an order of magnitude faster speed. Details are discussed below.

\textbf{Boundary Recall (BR).} 
Boundary recall is a measurement which denotes the adherence to
the boundaries.
It computes what fraction of the ground truth edges falls
within $\varepsilon$-pixel length from at least one
superpixel boundary.
The BR \cite{achanta2012slic} can be computed by
\begin{equation}\label{13}
BR_\mathcal{G}(\mathcal{S})=\frac{\sum_{p\in\xi_\mathcal{G}}\Pi
(\min_{q\in\xi_\mathcal{S}}\Vert{p-q}\Vert <\varepsilon)}{|\xi_\mathcal{G}|},
\end{equation}
where $\xi_\mathcal{S}$ and $\xi_\mathcal{G}$ respectively denote the union set of superpixel boundaries and the union set of ground truth boundaries.
The indicator function $\Pi$ checks
if the nearest pixel is within $\varepsilon$ distance. 
Here we follow \cite{achanta2012slic,liu2016manifold}
and set $\varepsilon = 2$ in our experiment.
The boundary recall curves of different methods are plotted in Figure \ref{Boundary recall}.
One can easily observe that our FLIC method outperforms all other methods.
	
\textbf{Undersegment Error (UE).} 
The undersegment error reflects the extent that superpixels do not exactly 
overlap the ground truth segmentation.
Similar to BR, UE can also reflect the boundary adherence.
The difference is that UE uses segmentation regions instead of boundaries
in the measurement.
Mathematically, the UE \cite{neubert2012superpixel} can be computed by
\begin{equation}\label{14}
UE_\mathcal{G}(\mathcal{S})=\frac{\sum_{G\in\varrho_\mathcal{G}}
			(\sum_{S:S\cap G \ne \phi} \min(S_{in},S_{out}))}{N},
\end{equation}	
where $\varrho_\mathcal{S}$ is the union set of superpixels,
$\varrho_\mathcal{G}$ is the union set of the segments of the ground truth,
$S_{in}$ denotes the overlapping of the superpixel $S$ and the ground truth segment $G$, and $S_{out}$ denotes the rest of the superpixel $S$.
As shown in Figure \ref{Undersegment error}, our results are nearly the same as those of
the best approach ERS~\cite{liu2011entropy} and run significantly faster. 

\begin{figure}[t]
\centering
	\subfigure[]{\includegraphics[width = 0.47\linewidth]{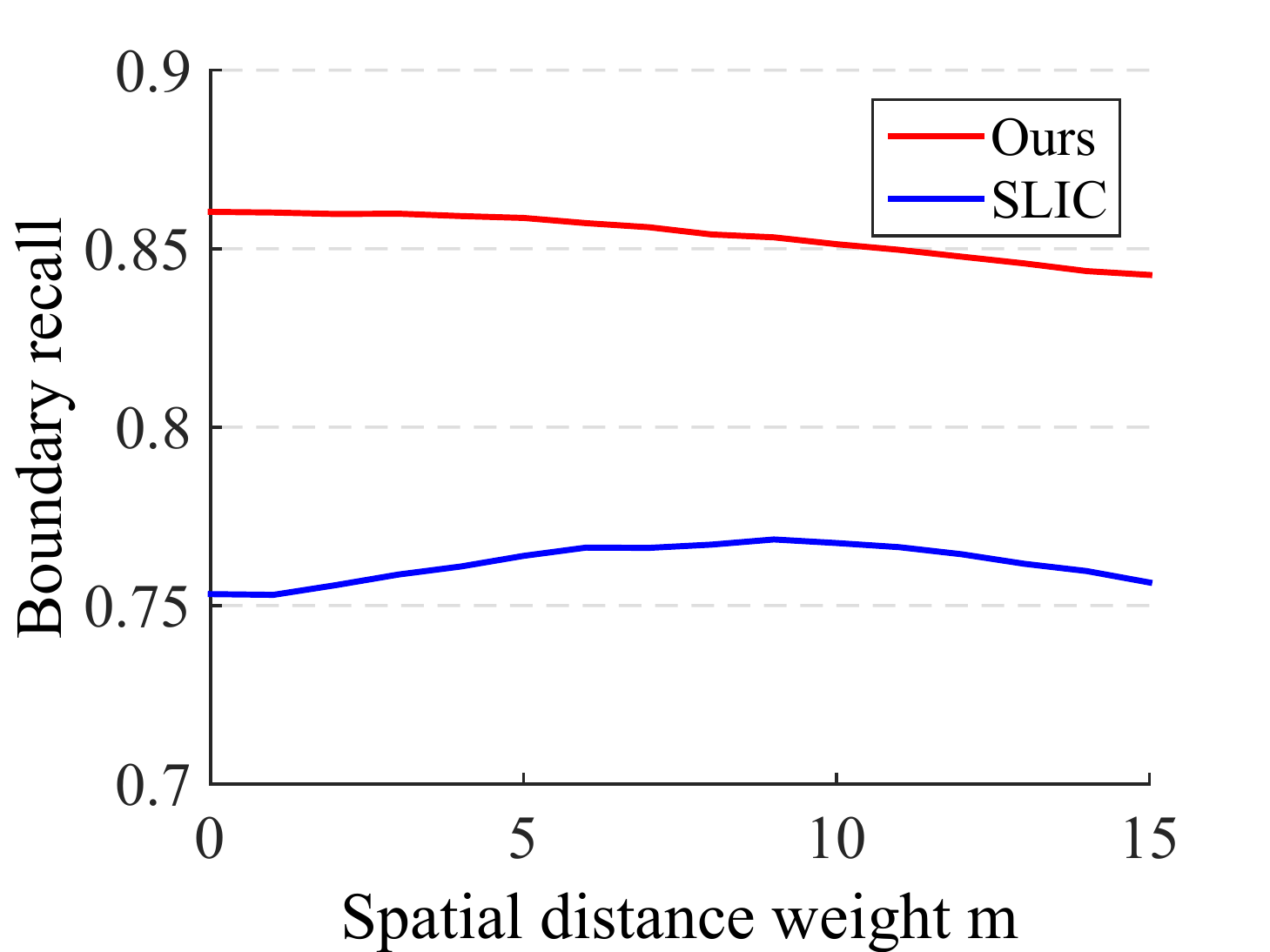}}
	\subfigure[]{\includegraphics[width = 0.47\linewidth]{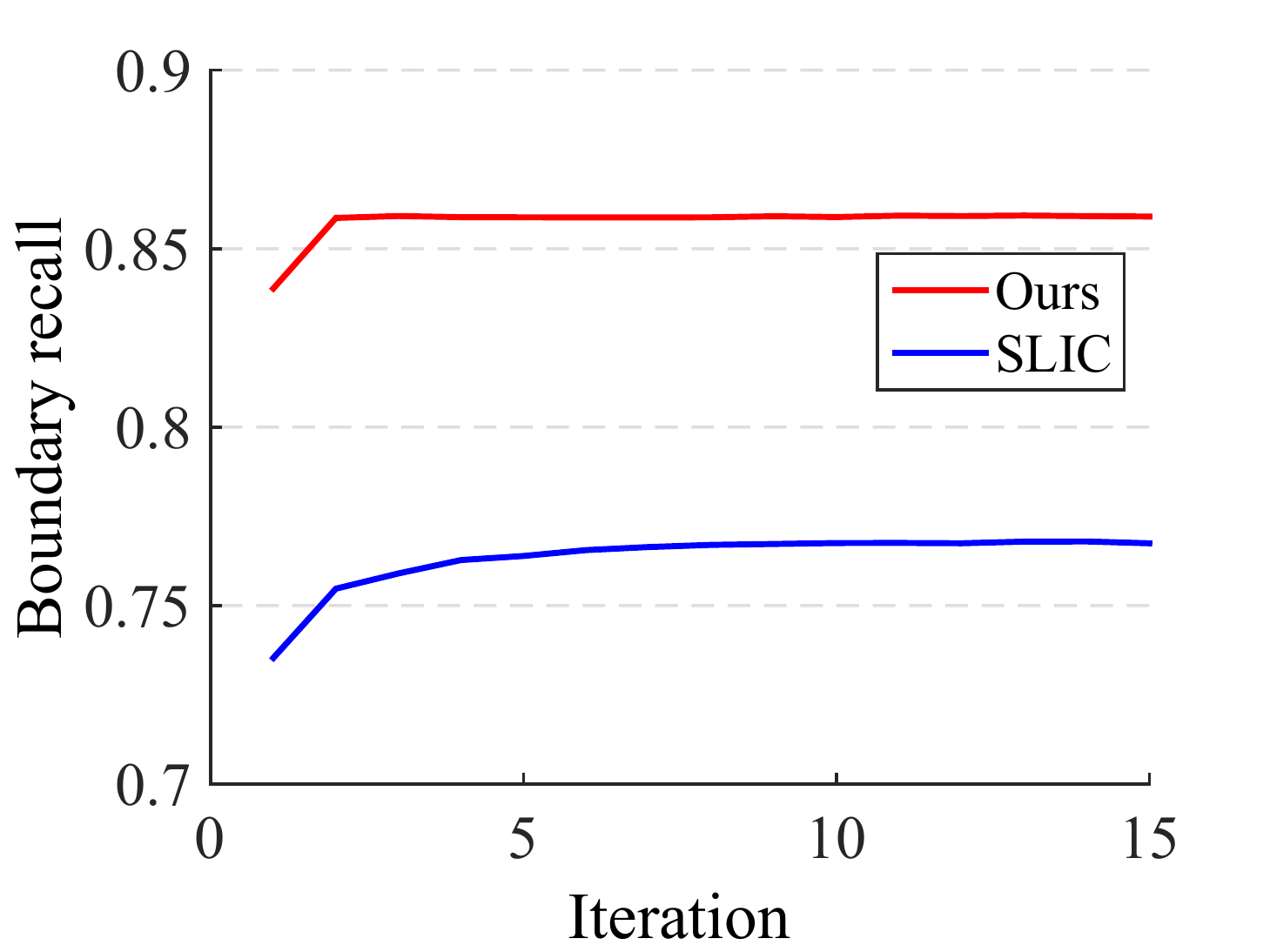}}
	\caption{(a)~The BR-$m$ curves, where $m$ is the
    spatial distance weight in Eqn. \ref{1}. Our overall performance is far
    better than SLIC for all the tested $m$.
    (b)~The BR-Iteration curves. Our method converges within 2 iterations,
    which is much faster than SLIC.
    }\label{para_influence}
\end{figure}

\begin{figure}[t]
	\begin{center}
		\includegraphics[width=0.85\linewidth]{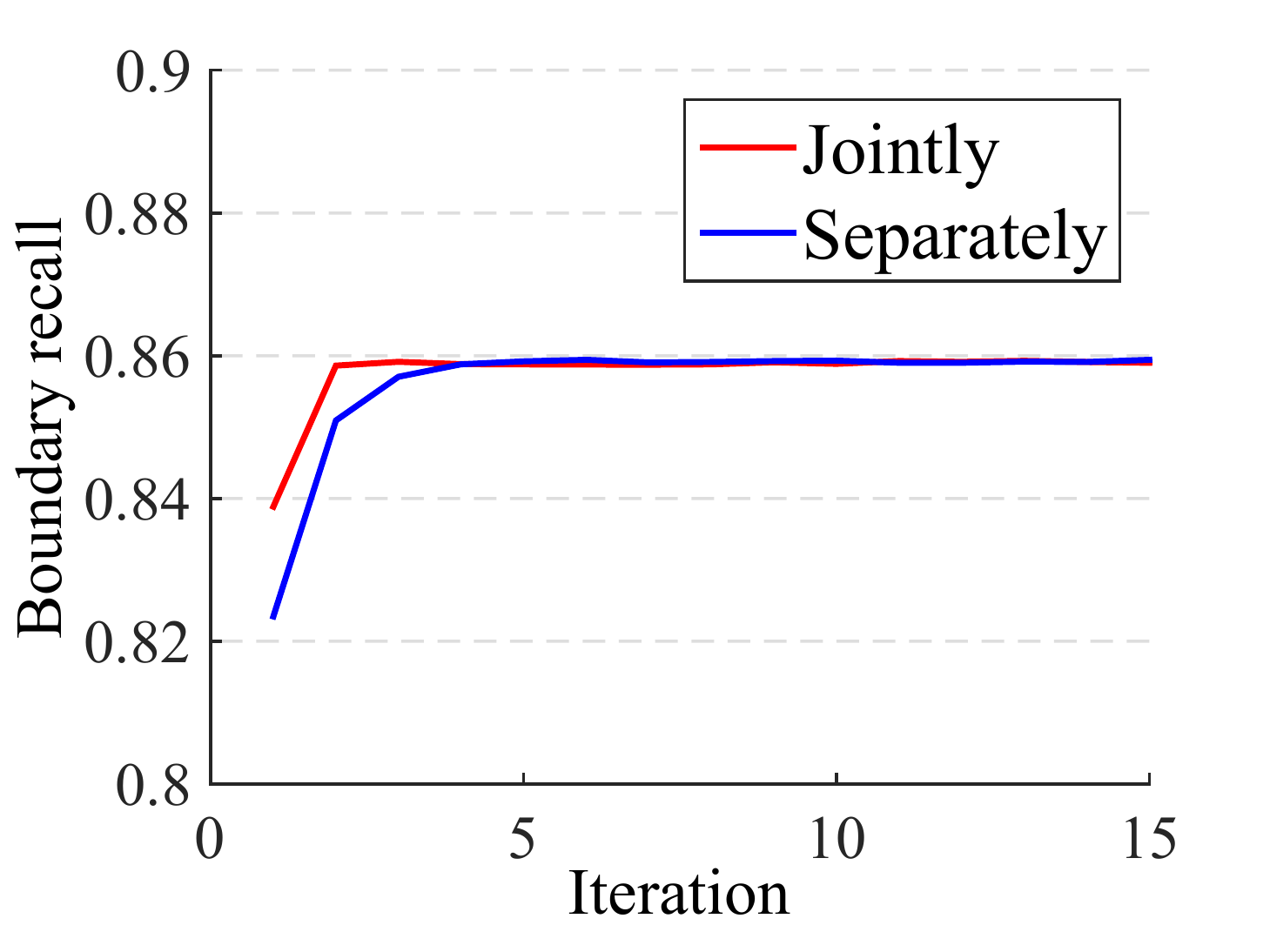}\\
		\caption{Comparison of the convergence rate between joint and
        separate assignment and update steps.
		} \label{fig:jointly}
	\end{center}
\end{figure}

\begin{table}[t] \title{ Boundary Recall(\%)}
	\centering
    \setlength\tabcolsep{4pt}
	\begin{tabular}{ccccccccc}\hline
      & \multicolumn{2}{c}{100} & \multicolumn{2}{c}{200} & \multicolumn{2}{c}{300} & \multicolumn{2}{c}{400} \\ \cline{2-3} \cline{4-5} \cline{6-7} \cline{8-9}
	  & BR & Time & BR & Time & BR & Time & BR & Time \\ \hline
	4-N & 78.6 & 34 & 85.9 & 35 & 89.1 & 36 & 91.8 & 38 \\
	8-N & 80.5 & 54 & 87.4 & 56 & 90.5 & 59 & 92.7 & 61\\ \hline
	\end{tabular}
    \caption{Boundary recall and time cost comparisons between 4-neighborhood and 8-neighborhood
    with different superpixel counts: 100, 200, 300, 400.}\label{BR_table}
\end{table}

\textbf{Achievable Segmentation Accuracy (ASA).}
ASA gives the highest accuracy achievable for object
segmentation that utilizes superpixels as units.
Similar to UE, ASA utilizes segments instead of the boundaries, which can be 
computed by \cite{liu2011entropy}
\begin{equation}\label{15}
		ASA_\mathcal{G}(\mathcal{S})=\frac{\sum_{k}
			\max_i|S_{k}\cap G_{i}|}
		{\sum_{i}G_i},
	\end{equation}
where $S_k$ represents the superpixel and $G_i$ represents the ground truth segment. A better superpixel segmentation will have a larger ASA value.
As shown in Figure \ref{fig:SegAcc}, compared to the ERS \cite{liu2011entropy}, the 
performance of our approach is competitive and our method achieves 
the best trade-off between the performance and time cost.
	
\textbf{Time Cost (TC).} 
Similar to SLIC, our method also achieves an $O(N)$ time complexity.
We know that computation efficiency is one of the most important points
for using superpixels as elementary units.
Many approaches are limited by their speed, such as SSS~\cite{wang2013structure} 
and ERS~\cite{liu2011entropy}.
As shown in Figure \ref{evaluation}, the average time cost of FLIC with
two iterations when processing an image is 0.035s, while the time costs of ERS,
Manifold SLIC, SLIC, and FH are 0.625, 0.281s, 0.072s, and 0.047s, respectively.
It is obvious that FLIC has the lowest time cost among all methods and it runs 
nearly 20 times faster than ERS with comparable result quality.

\begin{figure*}[t!]
  	\centering
  	\centering
    \setlength\tabcolsep{1.2pt}
    \begin{tabular}{cccccc}
		\includegraphics[width=0.16\linewidth]{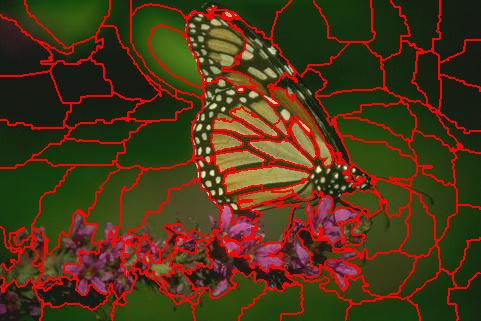} &
        \includegraphics[width=0.16\linewidth]{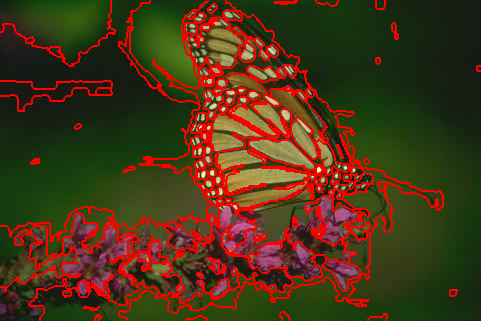} &
        \includegraphics[width=0.16\linewidth]{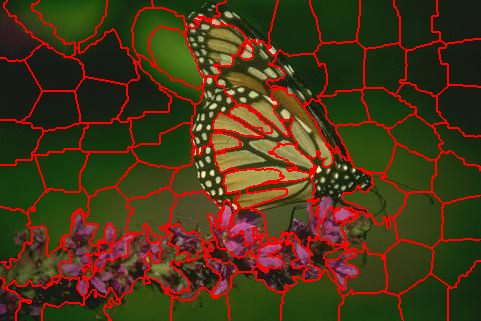} &
        \includegraphics[width=0.16\linewidth]{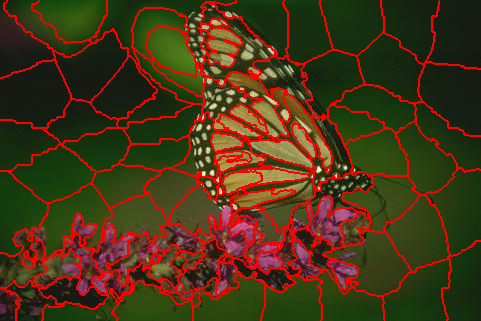} &
        \includegraphics[width=0.16\linewidth]{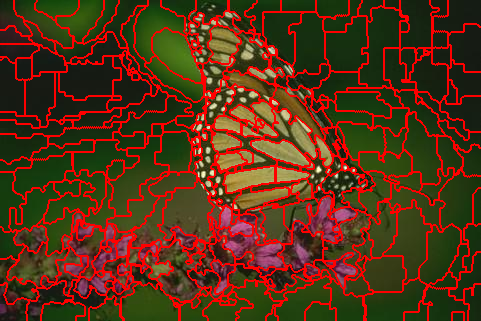} &
        \includegraphics[width=0.16\linewidth]{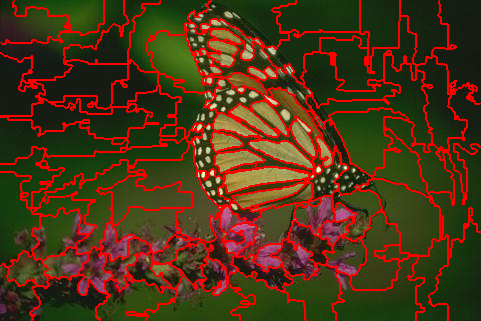} \\
  (a) Ours & (b) FH  & (c) SLIC & (d) M-SLIC  & (e) SEEDS & (f) ERS 
  \end{tabular}
  \caption{Visual comparison of superpixel segmentation results using different
  existing algorithm with 100 superpixels and $m = 10$. Our approach adheres to
  boundaries very well and at the same time produces compact superpixels.
  }\label{fig:visual}
\end{figure*}

\textbf{Visual Results and Analysis.}
In Figure \ref{fig:visual}, we show several superpixel segmentation 
results using different algorithms. 
As can be seen, our approach is more sensitive to image boundaries,
especially when there is poor contrast between the foreground and
background.
Compared to SLIC method, our approach adheres to boundaries very well 
and runs twice as fast.
Compared to ERS method, our resulting superpixels are much more regular 
and the mean execution time of our approach is 20 times shorter.

All the above facts and Figure \ref{evaluation} reflect that 
our approach achieves an excellent compromise among \emph{adherence},
\emph{compactness}, and \emph{time cost}.

\subsection{Algorithm Analysis}\label{sec:alg_analysis}

\begin{figure*}[t!]
  \centering
  \includegraphics[width=.33\linewidth]{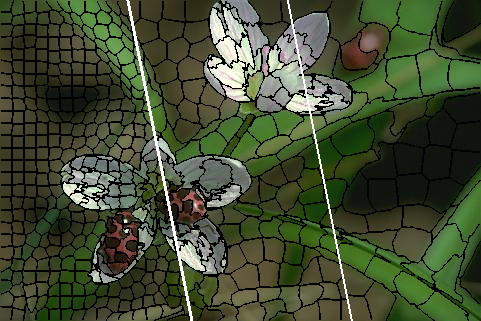}
  \includegraphics[width=.33\linewidth]{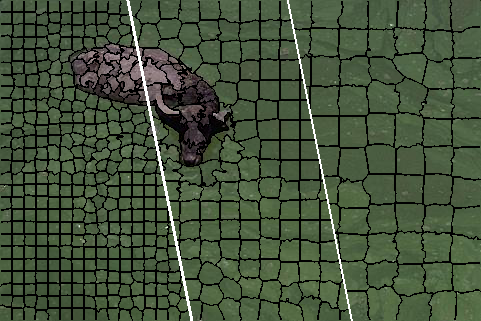}
  \includegraphics[width=.33\linewidth]{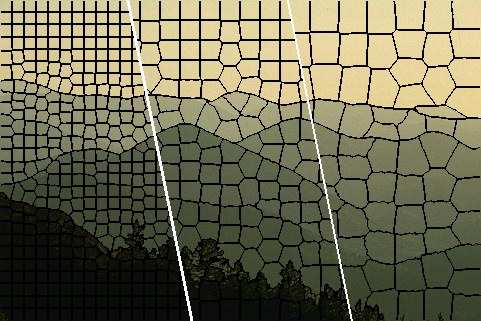}
  \caption{Images segmented by our proposed approach with $m=20$ and the number 
  	of superpixels set to 1000, 400, and 200, respectively. The resulting superpixels
    adhere to region boundaries very well.
  }\label{fig:multiscale_results}
\end{figure*}

\begin{figure*}[t!]
  \centering
  \includegraphics[width=.33\linewidth]{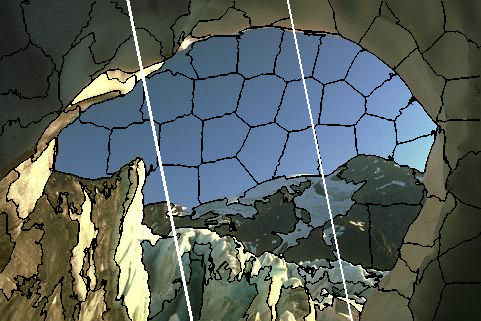}
  \includegraphics[width=.33\linewidth]{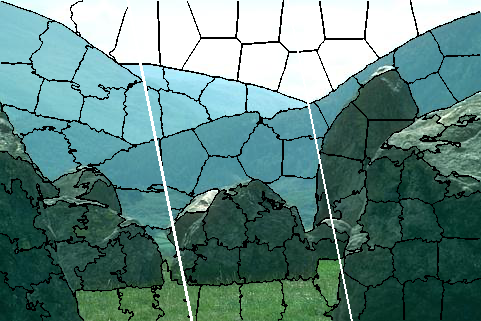}
  \includegraphics[width=.33\linewidth]{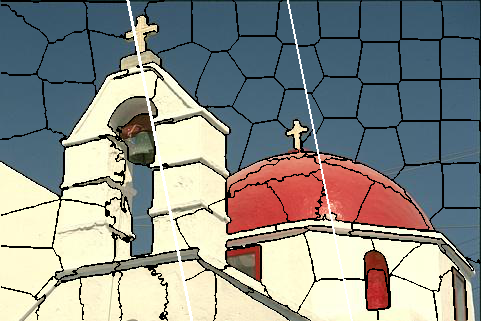}
  \caption{Images segmented by our proposed approach with $m=10,20$, and 30,
  respectively. When $m$ tends to be a smaller value, the superpixels
  adhere well to boundaries. When $m$ becomes larger, the superpixels become
  more compact.
  }\label{fig:m_results}
\end{figure*}

\textbf{Efficacy of the back-and-forth Traverse Order.}
As shown in Figure \ref{scan_normal}, we adopt a back-and-forth traverse order to
scan the whole region enclosed by a bounding box for each superpixel.
Actually, a couple of forward scans can also perform very well for our method.
We provide a comparison between two strategies: using pure forward scan order for four iterations versus using the proposed back-and-forth scan order twice (which is also four iterations).
Figure \ref{scan_order} shows quantitative comparisons between
these two strategies.
The blue line represents the results using normal forward scan order while
the red line stands for the results using our method.
It can be seen that the red curve significantly outperforms the blue 
one and achieves competitive time cost compared to the blue curve.
This fact reflects that our back-and-forth scan order considers more
information about the regions outside the bounding box, leading to more
reliable boundaries.

\textbf{The Role of the Spatial Distance Weight.}
As shown in Figure \ref{para_influence}(a), unlike SLIC \cite{achanta2012slic},
the BR curve with respect to the spatial distance weight $m$ is
monotonically decreasing in our approach.
The reason for this phenomenon is that in our method local region continuity is mostly ensured by the active search algorithm, and color boundaries are less well preserved for larger $m$. 
On the other hand, small $m$ will result in less regular superpixels, so we choose $m=5$ for our comparison with previous works.
It is noteworthy to mention that superpixels are normally considered as the
first step of most vision tasks and these vision tasks often favor those
superpixel methods with good boundaries.
Therefore, users can select a reasonable value for $m$ according to their specific
conditions.
In any case, our overall performance is significantly better for all $m$ values.

\textbf{Convergence Rate.} 
FLIC significantly accelerates the evolution so that we only need a few
iterations before convergence. 
We compare the performance curves with different iterations
on the Berkeley benchmark. 
It can be easily found in Figure \ref{para_influence}(b) that our algorithm quickly
converges within only 
two iterations and more iterations only bring marginal benefits to the results.
Numerically, the boundary recall of the superpixels with only one iteration is
0.835 when $K$ is set to 200.
The value after two iterations is 0.859 and after three iterations it is 0.860 
when generating the same number of superpixels.
The undersegment error values are 0.115, 0.108, and 0.107, respectively.
The achievable segmentation accuracy values are 0.941, 0.945, and 0.946, respectively.
As can be seen in Figure \ref{para_influence}(b), our algorithm not only converges much 
faster than SLIC (which requires ten iterations to converge), but also obtains
better performance.

\textbf{The Role of the Joint Assignment and Update}.
Our algorithm jointly performs the assignment and update steps. 
In Figure \ref{fig:jointly}, we show the convergence rates of both our joint approach 
and that of separately performing assignment and update steps. 
One can observe that our joint approach converges very quickly and only
two iterations are needed, while the separate approach needs another two iterations
to reach the same BR value.
This phenomenon demonstrates that our joint approach is efficient and without
any negative effect on our final results.

\textbf{Effect on the Size of Neighborhoods}. 
As mentioned in Section \ref{sec:method}, in our implementation, the label of the current
pixel relies on its four neighborhood pixels. 
Actually, using eight neighborhood pixels is also reasonable as more neighbors
will definitely provide more useful information.
In Table \ref{BR_table}, we briefly compare the results
for these two cases. 
A natural observation is that using larger neighborhoods leads to an increase in performance but
at the cost of reducing running speed.
With regard to real applications, users can select either case to suit their own preferences.
    
\textbf{Qualitative Results.}
In Figure \ref{fig:multiscale_results} we show some segmentation results
produced by our approach with $m=20$ and the number of superpixels set 
to 1000, 400, and 200, respectively.
It is seen that, over the range of $K$ value, the edges of the resulting
superpixels are always very close to the boundaries.
This phenomenon is especially obvious in the first image and the third
image.
We also show some segmentation results with different values of $m$ in 
Figure \ref{fig:m_results}.
When $m$ tends to smaller values, for example 10, the shapes of the 
resulting superpixels become less regular.
When $m$ is larger, for example 30, the resulting superpixels become more
compact.

\section{Conclusions} \label{sec:conclusion}
In this paper we present a novel algorithm using active search,
which is able to improve the performance and
significantly reduce the time cost for using superpixels to over-segment an image.
Taking advantage of local continuity,
our algorithm provides results with good boundary sensitivity
even for complex and low contrast image.
Moreover, it is able to converge in only two iterations,
achieving the lowest time cost compared to previous methods
while obtaining performance comparable to the state-of-the-art
method ERS with 1/20th of its running time.
We have used various evaluation metrics on the Berkeley segmentation benchmark dataset
to demonstrate the high efficiency and high performance of our approach.

\section*{Acknowledgments}
This research was sponsored by NSFC (61620106008, 61572264),
CAST (YESS20150117),
Huawei Innovation Research Program (HIRP),
and IBM Global SUR award.

{\small
\bibliographystyle{aaai}
\bibliography{FLIC_AAAI}
}

\end{document}